\documentclass[10pt,twocolumn,letterpaper]{article}

\usepackage{iccv}
\usepackage{times}
\usepackage{epsfig}
\usepackage{graphicx}
\usepackage{amsmath}
\usepackage{amssymb}
\usepackage{multirow}

\usepackage[top=2cm, bottom=2cm, left=2cm, right=2cm]{geometry}
\usepackage{algorithm}
\usepackage{algorithmicx}
\usepackage{algpseudocode}
\usepackage{amsmath}
\usepackage{threeparttable}
\usepackage[breaklinks=true,bookmarks=false]{hyperref}

\iccvfinalcopy % *** Uncomment this line for the final submission

\def\iccvPaperID{9006} % *** Enter the ICCV Paper ID here

\ificcvfinal\fi

\begin{document}

%\bibliographystyle{plain}

%%%%%%%%% TITLE
\title{Control Distance IoU and Control Distance IoU Loss Function for Better Bounding Box Regression}

\author{Chen Dong\\
Tongji University\\
No.4800, Cao'an Highway, Jiading District, Shanghai\\
{\tt\small 1910691@tongji.edu.cn}
\and
Miao Duoqian\\
Tongji University\\
No.4800, Cao'an Highway, Jiading District, Shanghai\\
{\tt\small dqmiao@tongji.edu.cn}
}

\maketitle
\ificcvfinal\thispagestyle{empty}\fi

%%%%%%%%% ABSTRACT%%%%%%%%%%%%%%%%%%%%%%%%%%%
\begin{abstract}
   Numerous improvements for feedback mechanisms have contributed to the great progress in object detection.
In this paper, we first present an evaluation-feedback module, which is proposed to consist of evaluation system and feedback mechanism.
Then we analyze and summarize the disadvantages and improvements of traditional evaluation-feedback module.
Finally, we focus on both the evaluation system and the feedback mechanism,
and propose \textbf{Control Distance IoU} and \textbf{Control Distance IoU loss function} (or \textbf{CDIoU} and \textbf{CDIoU loss} for short)
without increasing parameters or FLOPs in models,
which show different significant enhancements on several classical and emerging models.
Some experiments and comparative tests show that coordinated
evaluation-feedback module can effectively improve model performance.
CDIoU and CDIoU loss have different excellent performances in several models such as Faster R-CNN, YOLOv4, RetinaNet and ATSS.
There is a maximum AP improvement of 1.9\% and an average AP of 0.8\% improvement on MS COCO dataset,
compared to traditional evaluation-feedback modules.
\end{abstract}

%%%%%%%%% BODY TEXT %%%%%%%%%%%%%%%%%%%%%%%%%
%-------------------------------------------------------------------------
\section{Introduction}

Tremendous works have been made for more accurate and more efficient object detection
in recent years. Data augmentation\cite{2019AutoAugment,kisantal2019augmentation,zoph2020learning},
deeper layers of neural networks\cite{fu2017dssd,liu2016ssd,2020EfficientDet},
more complex structured FPN modules\cite{li2017fssd,lin2017feature,2020EfficientDet},
and even more number of iterations make the model for object detection state-of-the-art. Undoubtedly these models
have achieved remarkable success, however, at the same time these models have huge parameters and unsatisfactory FLOPs,
such as Detectron2 Mask R-CNN X101-FPN (parameters: 107M, FLOPs: 277B)\cite{wu2019detectron2}, ResNet-50 + NAS-FPN (1280@384)
(parameters: 104M, FLOPs: 1043B)\cite{2019NAS}, AmoebaNet+ NAS-FPN +AA(1280) (parameters: 185M, FLOPs: 1317B)\cite{zoph2020learning} and
AmoebaNet+ NAS-FPN + AA(1536) (parameters: 209M, FLOPs: 3045B)\cite{zoph2020learning}.

This paper focuses the performance improvement of object detection on the \textbf{evaluation system}
and \textbf{feedback mechanism}\cite{wen2016discriminative} (namely IoU modules and loss functions, combined called \textbf{evaluation-feedback module}) of region proposals
without increasing the number of parameters or FLOPs.

The evaluation-feedback modules have 3 main roles:
(1) Evaluating region proposals, using ground truth as a criterion.
(2) Ranking a set of region proposals (with the same ground truth criterion).
(3) Feeding the gap between region proposals (RP) and ground truths (GT) to the neural network,
which is used to correct the next evaluation module. Considering evaluation-feedback module is fundamental,
this module should be efficient and contain few parameters.
A good evaluation-feedback module should meet the following 3 conditions:
\begin{itemize}
\setlength{\itemsep}{0pt}
\setlength{\parsep}{0pt}
\setlength{\parskip}{0pt}
  \item A measure overlapping area.
  \item The good ability to differentiate and a measure of the degree of difference
(sometimes understood as centroid distance and aspect ratio).
  \item The IoUs calculation can be correlated with loss functions calculation.
\end{itemize}

Numerous previous studies have tended to focus on the study of feedback mechanism
at the expense of evaluation system. In this paper, the Control Distance IoU(\textbf{CDIoU} for short)
and the Control Distance IoU loss function (\textbf{CDIoU loss} for short) are proposed and given the same importance.
CDIoU has good continuity and derivability,
and simplifies the calculation by measuring the distance
between RP and GT in a unified way,
optimizing the calculation of DIoU and CIoU\cite{2020Distance} for centroid distance and aspect ratio,
and completing the evaluation quickly.
The CDIoU loss function can be correlated with CDIoU calculation,
which enables the feedback mechanism to characterize more accurately and
feed back the difference between RP and GT,
thus making the objective function of the deep learning network converge faster
and improving the overall efficiency. The CDIoU and CDIoU loss functions
are highly adaptive and show significant improvements on several different models,
compared to traditional IoU modules and loss functions.
 The main contributions of this work can be summarized as:
 \begin{itemize}
\setlength{\itemsep}{0pt}
\setlength{\parsep}{0pt}
\setlength{\parskip}{0pt}
  \item The \textbf{evaluation-feedback module} is proposed to consist of \textbf{evaluation system} and \textbf{feedback mechanism}.
  \item \textbf{CDIoU} is proposed as a new evaluation system and \textbf{CDIoU loss} as a new feedback mechanism.
  \item Improving the results, while the number of parameters and running time are not increased.
  \item With wide applicability, make significant improvements on several models.
\end{itemize}
%------------------------------------------------------------------------
\section{Related Work}

The first culmination of deep learning for object detection was the proposal of R-CNN\cite{2013Rich},
Fast R-CNN\cite{2015Fast} and Faster R-CNN\cite{2017Faster} models, which laid down the basic framework and data processing for deep learning applied to object detection. YOLO\cite{redmon2016you} provides a more straightforward way by directly regressing the location of the bounding box and the class to which the
bounding exploitation belongs in the output layer, thus transforming the object detection
problem into a regression problem. After this, YOLOv2\cite{2017YOLO9000}, YOLOv3\cite{farhadi2018yolov3},
YOLOv4\cite{bochkovskiy2020yolov4} and YOLOv5\cite{glenn_jocher_2021_4418161} were proposed,
which made the deep learning network not only improve in accuracy but also in computing speed.
R-CNN series and YOLO series are the classical representatives of two-stage model\cite{dai2016r,duan2020corner,he2017mask,he2015spatial,he2019multi} and
one-stage model\cite{2020FCOS,wu2017squeezedet} in object detection.

\textbf{Neural network backbone and conv kernel}

The backbone networks of deep learning are also evolving. LeNet (1998)\cite{1998Gradient},
AlexNet (2012)\cite{krizhevsky2017imagenet}, VGGNet (2014)\cite{simonyan2014very},
GoogLeNet (2014)\cite{2014Going}, ResNet (2015)\cite{2016Deep}, and MobileNet (2017)\cite{howard2017mobilenets}
are preserved in the path of deep learning development.
EfficientNet (2019)\cite{tan2019efficientnet} proposes a more generalized idea on the optimization of current
classification networks, arguing that the three common ways of enhancing network metrics,
namely widening the network, deepening the network and increasing the resolution,
should not be independent of each other.

Along with the backbone, the convolutional kernel\cite{fu2017dssd,noh2015learning,radford2015unsupervised} is also evolving and changing.
Deformable conv\cite{dai2017deformable,2019Deformable} adds an offset variable to the position of each sampled point in the convolution kernel,
enabling random sampling around the current position without being restricted to the previous regular grid points.
Dilated conv\cite{chen2017rethinking,2016Multi} can effectively focus on the semantic information of the local pixel blocks,
instead of letting each pixel rub together with the surrounding blocks, which affects the detail of segmentation.

\textbf{Evaluation-feedback module}

Based on IoU, GIoU\cite{2019Generalized} focuses not only on overlapping regions but also on other non-overlapping regions,
which can better reflect the overlap of RP and GT.
DIoU\cite{2020Distance} takes the distance between object and anchor, overlap rate and scale into consideration,
which makes the object box regression more stable and does not have problems such as scattering during training like IoU and GIoU.

GIoU loss\cite{2016Multi} still has the problems of slow convergence and inaccurate regression.
It is found that GIoU first tries to overlap the object box by increasing the size of the detection box,
and then uses the IoU loss term to maximize the overlap area with the object box.

DIoU loss and CIoU loss\cite{2020Distance} greatly enriched the connotation of IoU calculation results,
adding the measurement of difference, including ``centroid distance"
and ``aspect ratio" separately. DIoU loss cannot distinguish which RPs is more similar to
GT when the center points of RPs are at the same position. We can know that the calculation process of CIoU loss is more time-consuming,
which will eventually drag down the overall training and test time.

%------------------------------------------------------------------------
\section{Analysis of traditional IoUs and loss functions}
In object detection, the function of IoUs is to
evaluate the similarity between RP and GT. Through the IoU method,
the evaluation between RP and GT is given,
which plays a fundamental role in the selection of positive and negative samples. In the
evaluation-feedback module, the most representative methods are IoU, GIoU, DIoU loss and CIoU loss.
They have played a fundamental role in the great progress of object detection,
but there still is much room for optimization.

\subsection{Analysis of traditional IoUs}
IoU is a basic evaluation method, and as shown in Figure 1,
the relative position relationship between RP and GT is obviously
different. The human brain can clearly distinguish the disadvantages,
but the evaluation results of IoU are the same.

\begin{figure}[h]
\centering
\includegraphics[scale=0.3]{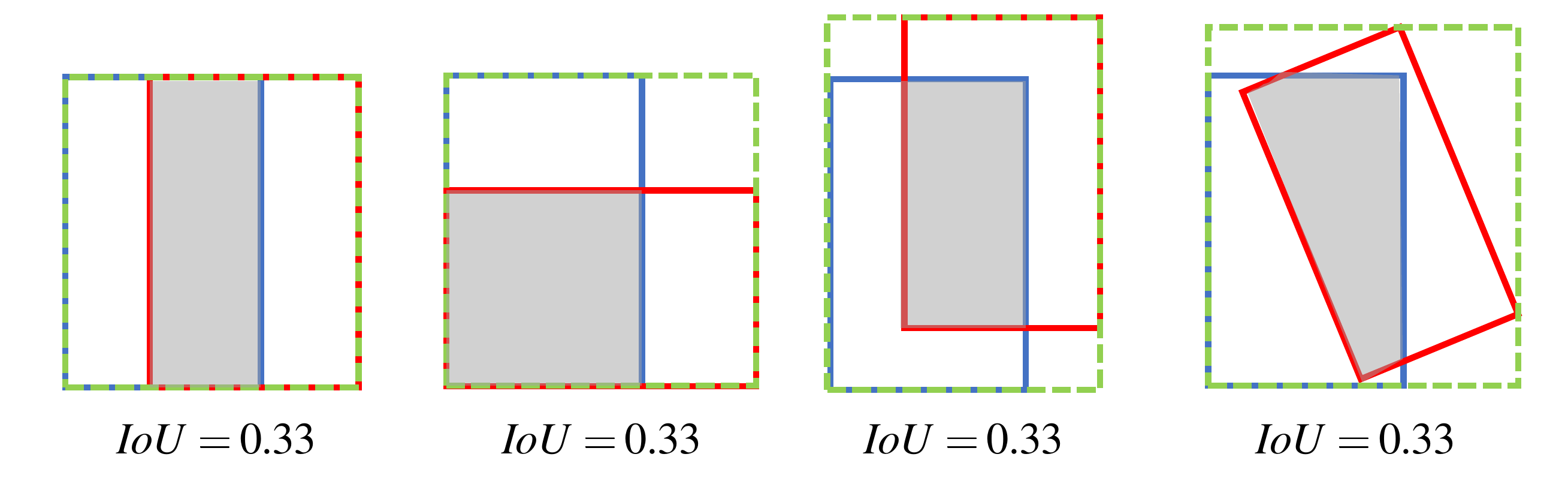}
\caption{Different relative position relations, the same IoU results}
\label{fig:label}
\end{figure}

Based on original IoU, many evaluation systems are derived, which enrich the evaluation
dimensions of previous IoU from those different aspects.

\begin{figure}[h]
\centering
\includegraphics[scale=0.5]{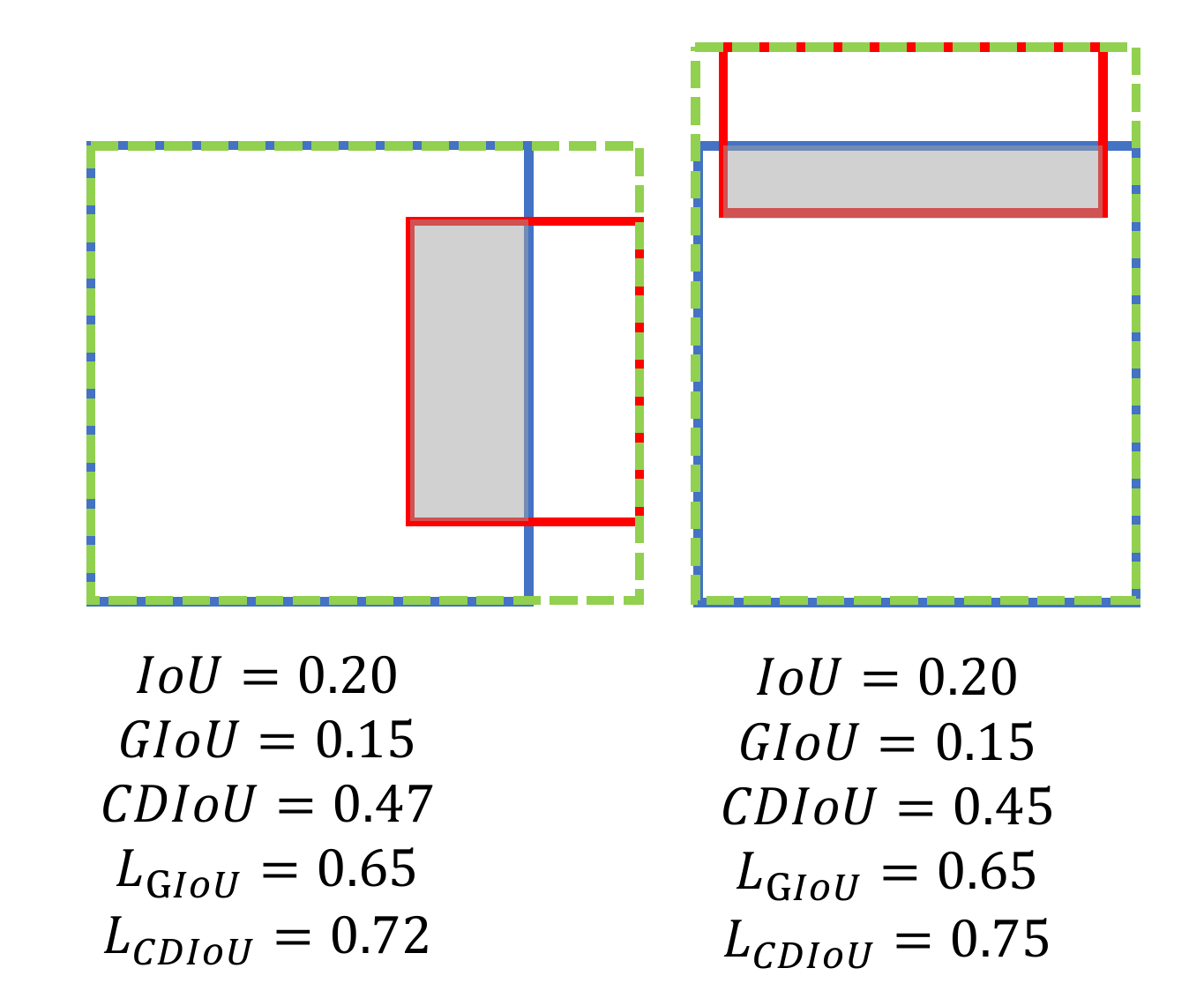}
\caption{Comparison of IoU GIoU CDIoU and Loss of GIoU \& CDIoU}
\label{fig:label}
\end{figure}

IoU only considers the calculation of overlapping area. Meanwhile, GIoU pays attention to
overlapping area and non overlapping area, and strengthens the discussion of evaluation system.
However, GIoU obviously ignored the ``measurement of difference" between RP and GT.

\begin{figure}[h]
\centering
\includegraphics[scale=0.3]{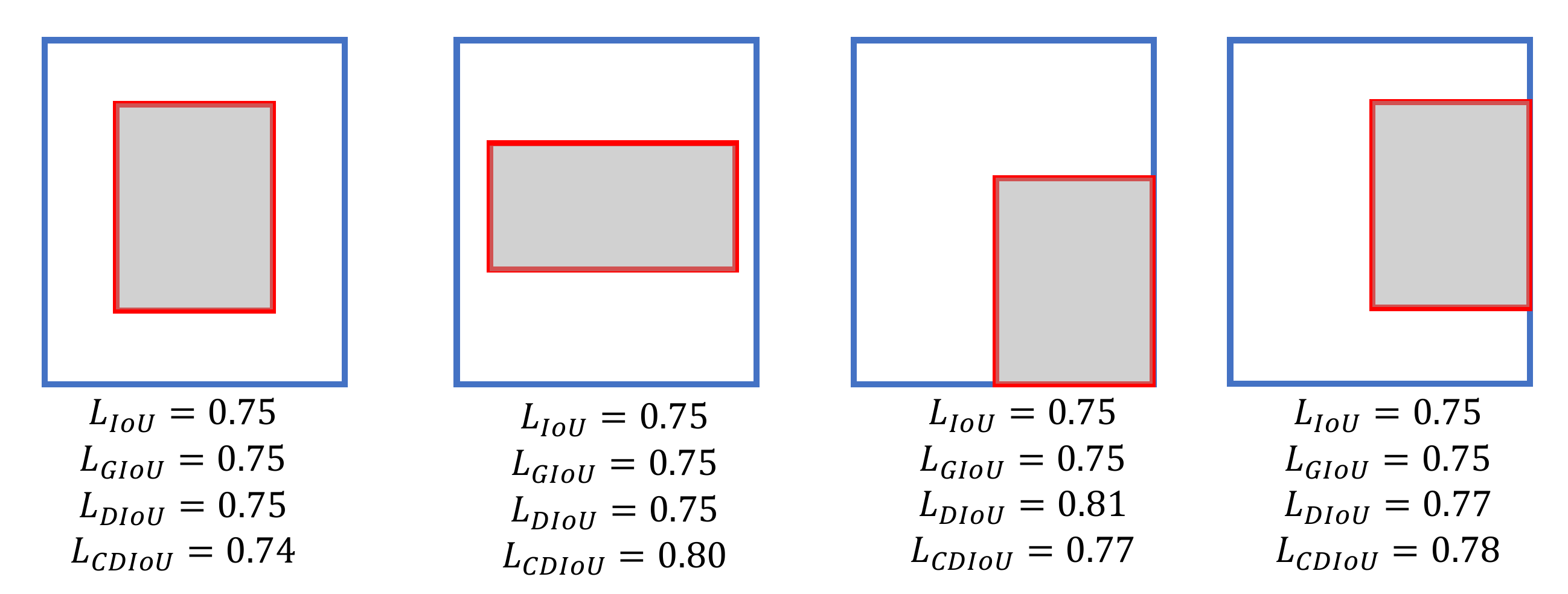}
\caption{Comparison of multiple IoU losses}
\label{fig:label}
\end{figure}

The ``measurement of difference" between RP and GT
include the distance between the center points (centroid) and the ratio of length-width(aspect ratio).
DIoU takes centroid into account in the calculation
of the evaluation system, but omits aspect ratio. As shown in Figure 2, 3,
DIoU can't recognize the difference between the high-thin region proposal and the short-fat region proposal,
and gives them the same value. But in fact, the human brain can easily differentiate which one is better and which position is better.

\subsection{IoU:Smooth L1 Loss and IoU Loss}

The method of smooth loss is proposed from Fast RCNN\cite{2015Fast},
which initially solves the problem of characterizing the boundary box loss.
Assuming that \begin{math}x\end{math} is the numerical difference between RP and GT,
\begin{math}L_{1}\end{math} and \begin{math}L_{2}\end{math}
loss are commonly defined as:\\
\begin{equation}
\mathcal{L}_{1}=|x| \frac{d \mathcal{L}_{2}(x)}{x}=2 x,
\end{equation}

\begin{equation}
\mathcal{L}_{2}=x^{2}.
\end{equation}
The corresponding derivative : 
\begin{equation}
\frac{d \mathcal{L}_{1}(x)}{x}=\left\{\begin{array}{ll}
 1 & \text{, if } x \geq 0 \\
-1 & \text {, otherswise ,}
\end{array}\right.
\end{equation}

\begin{equation}
\frac{d \mathcal{L}_{2}(x)}{x}=2 x.
\end{equation}

From the derivative of loss function to \begin{math}x\end{math}, we can know that the derivative of
loss function \begin{math}\mathcal{L}_{1}\end{math} to \begin{math}x\end{math} is constant. In the late training period,
when \begin{math}x\end{math} is very small, if the learning rate is constant, the loss function will fluctuate
around the stable value, and it is difficult to converge to higher accuracy. When the
derivative of loss function \begin{math}\mathcal{L}_{2}\end{math} to \begin{math}x\end{math} is large, its derivative
is also very large and unstable at the beginning of training. \begin{math}{smooth}_{\mathcal{L}_{1}}(x)\end{math} perfectly
avoids the shortcomings of \begin{math}\mathcal{L}_{1}\end{math} and \begin{math}\mathcal{L}_{2}\end{math} loss.

\begin{equation}
\text {smooth}_{\mathcal{L}_{1}}(x)=\left\{\begin{array}{ll}
0.5 x^{2} & \text {, if }|x|<1 \\
|x|-0.5 & \text {, otherswise   ,}
\end{array}\right.
\end{equation}

\begin{equation}
\frac{d {smooth}_{\mathcal{L}_{1}(x)}}{x}=\left\{\begin{array}{ll}
x & \text{, if }|x|<1 \\
\pm 1 & \text {, otherswise   ,}
\end{array}\right.
\end{equation}

However, in the actual object detection, the loss in box regression task is

\begin{equation}
\mathcal{L}_{l o c}\left(t^{u}, v\right)=\sum_{i \in\{x, y, w, h\}} \text {smooth}_{\mathcal{L}_{1}}\left(t_{i}^{u}-v_{i}\right),
\end{equation}

Where \begin{math}
   v=\left(v_{x}, v_{y}, v_{w}, v_{h}\right)
\end{math}represents the box coordinates of GT, and \begin{math}
   t^{u}=\left(t_{x}^{u}, t_{y}^{u}, t_{w}^{u}, t_{h}^{u}\right)
   \end{math} represents the predicted box coordinates,
that is to calculate the loss of four points respectively,
and then add them as the bounding box regression loss.

\textbf{shortcomings:}

1.When the above losses are used to calculate the bounding box loss of object detection,
the loss of four points is calculated independently, and then the final bounding box loss is obtained by adding.
The assumption of this method is that the four points are independent of each other,
and there is a certain correlation in fact. The calculation of smooth can not be unified with IoU, which leads to errors of the feedback mechanism and evaluation system.

2.The actual indicator of evaluation is to use IoU, which is not equivalent.
IoU loss cannot avoid this scenario,``different RPs, same feedback results'' in Figure 2,3.

\begin{equation}
\mathcal{L}_{IoU}=-\ln(IoU),
\end{equation}

\begin{equation}
\mathcal{L}_{IoU}=1-IoU.
\end{equation}

\subsection{GIoU and GIoU Loss}
On the premise of not increasing the calculation time,
GIoU\cite{2019Generalized} initially optimized the calculation of IoU for overlapping area,
and reduced the calculation error, but GIoU still did not take the measurement
of the difference into account in the calculation results.
\begin{equation}
G I o U=I o U-\frac{|C \backslash(A \cup B)|}{|C|},
\end{equation}

\begin{equation}
\mathcal{L}_{G I o U}=1-G I o U .
\end{equation}

GIoU loss still has the problems of slow convergence and inaccurate regression.
It is found that GIoU first tries to overlap the object box (GT) by increasing size of the detection box (RP),
and then uses IoU loss term to maximize the overlap area with the object box. At the same time,
when the two boxes contain each other, GIoU loss will degenerate
into IoU loss. At this time, the alignment of the bounding box becomes more difficult and the convergence is slow.

\subsection{DIoU loss and CIoU Loss}

DIoU loss and CIoU loss\cite{2020Distance} greatly enriched the connotation of IoU calculation results,
adding the measurement of difference, including ``centroid distance"
and ``aspect ratio" separately.

\begin{equation}
\mathcal{L}_{D I o U}=1-I o U+\frac{\rho^{2}\left(\mathbf{b}, \mathbf{b}^{g t}\right)}{c^{2}}
\end{equation}

First of all, DIoU loss cannot distinguish which region proposals are more similar to
 ground truth when the center points of region
proposals are at the same position. Then when two boxes are completely coincident,
\begin{math}
\mathcal{L}_{I o U}=\mathcal{L}_{G I o U}=\mathcal{L}_{D I o U}=0
\end{math}; when two boxes do not intersect, GIoU loss can't distinguish region proposals exactly and \begin{math}
\mathcal{L}_{G I o U}=\mathcal{L}_{D I o U} \rightarrow 2
\end{math}.

As shown in the Figure 4, in fact, as long as the center point of the region proposal is on arc \begin{math}
C\end{math} of circle \begin{math}O\end{math},the penalty terms of DIoU loss are consistent.
This is DIoU, which loses the accuracy of the evaluation system.

\begin{figure}[h]
\centering
\includegraphics[scale=0.34]{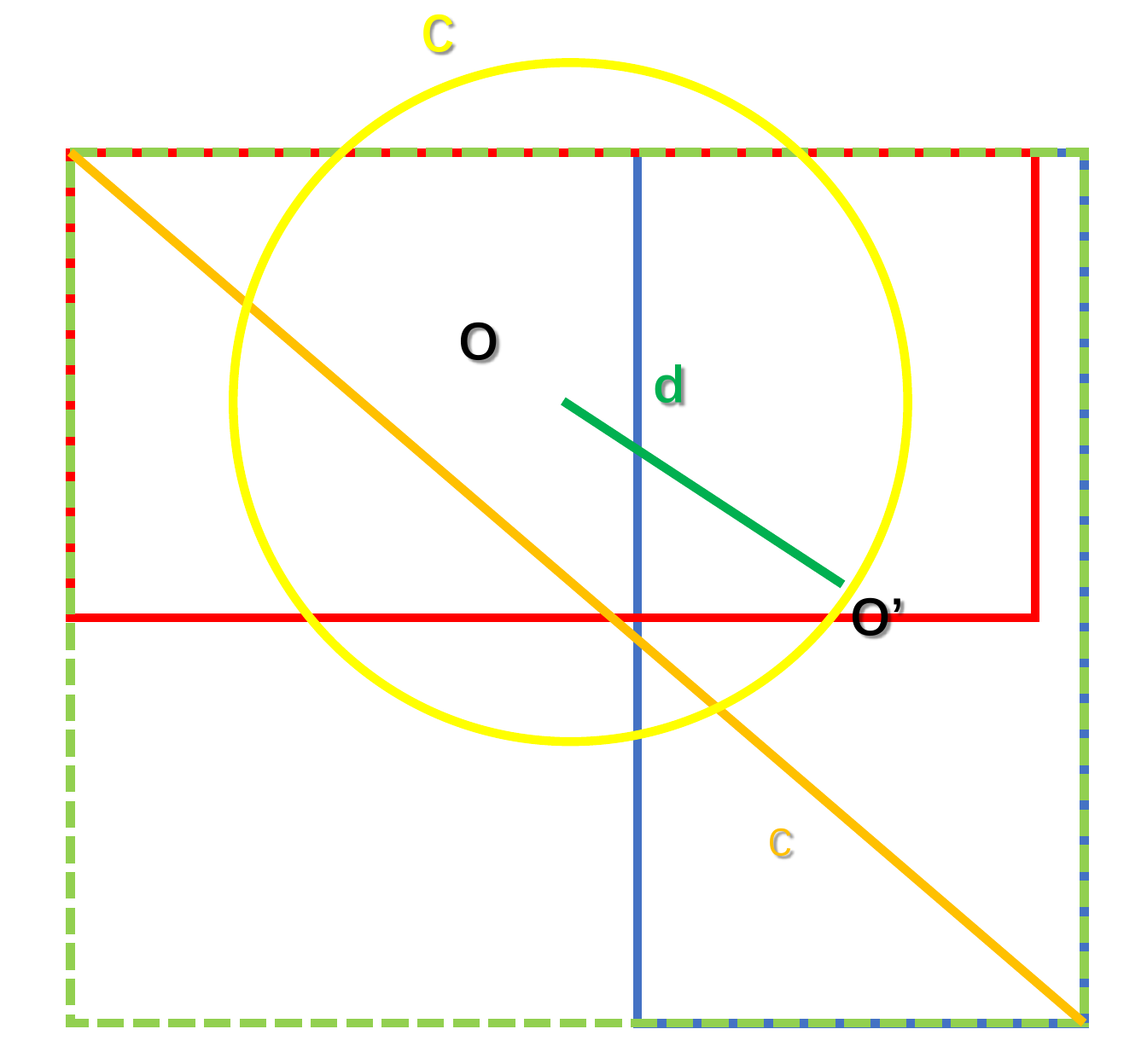}
\caption{DIoU loss: unuseful relative position relationship between RP and GT}
\label{fig:label}
\end{figure}

The penalty term of CIoU loss is composed of a factor \begin{math}
\alpha v
\end{math} and DIoU loss penalty term,
which takes that into account the aspect ratio of RP and GT.
\begin{equation}
\mathcal{L}_{C I o U}=1-I o U+\frac{\rho^{2}\left(\mathbf{b}, \mathbf{b}^{g t}\right)}{c^{2}}+\alpha v
\end{equation}

The penalty is

\begin{equation}
R_{C I o U}=\frac{\rho^{2}\left(b, b^{g t}\right)}{c^{2}}+\alpha v,
\end{equation}

\begin{equation}
\alpha=\frac{v}{(1-I o U)+v},
\end{equation}

\begin{equation}
v=\frac{4}{\pi^{2}}\left(\arctan \frac{w^{g t}}{h^{g t}}-\arctan \frac{w}{h}\right)^{2}.
\end{equation}

Because the calculation of CIoU loss involves the inverse trigonometric function,
and through comparative experiments, we can know that the calculation process of CIoU loss is more time-consuming,
which will eventually drag down the overall training time. For detailed comparison tests, see ``\textbf{Ablation studies}".

\section{CDIoU and CDIoU loss functions}

Based on  traditional IoUs and loss functions, CDIoU and CDIoU loss functions are proposed in this paper. Without increasing the operation time,
the running efficiency and AP are significantly improved. The CDIoU loss
function converges faster and reduces the complexity of the operation significantly.

Control Distance Intersection over Union (CDIoU) is an evaluation method that directly examines the similarity of RP and GT,
and it does not directly measure the distance between their centroids and the similarity of their shapes. For detailed information, see Figure 5.

\begin{equation}
\begin{aligned}
\operatorname{diou} &=\frac{\|R P-G T\|_{2}}{4\mathrm{MBR}^{\prime} s \text { diagonal }} \\
&=\frac{A E+B F+C G+D H}{4 W Y}
\end{aligned}
\end{equation}

\begin{figure}[h]
\centering
\includegraphics[scale=0.22]{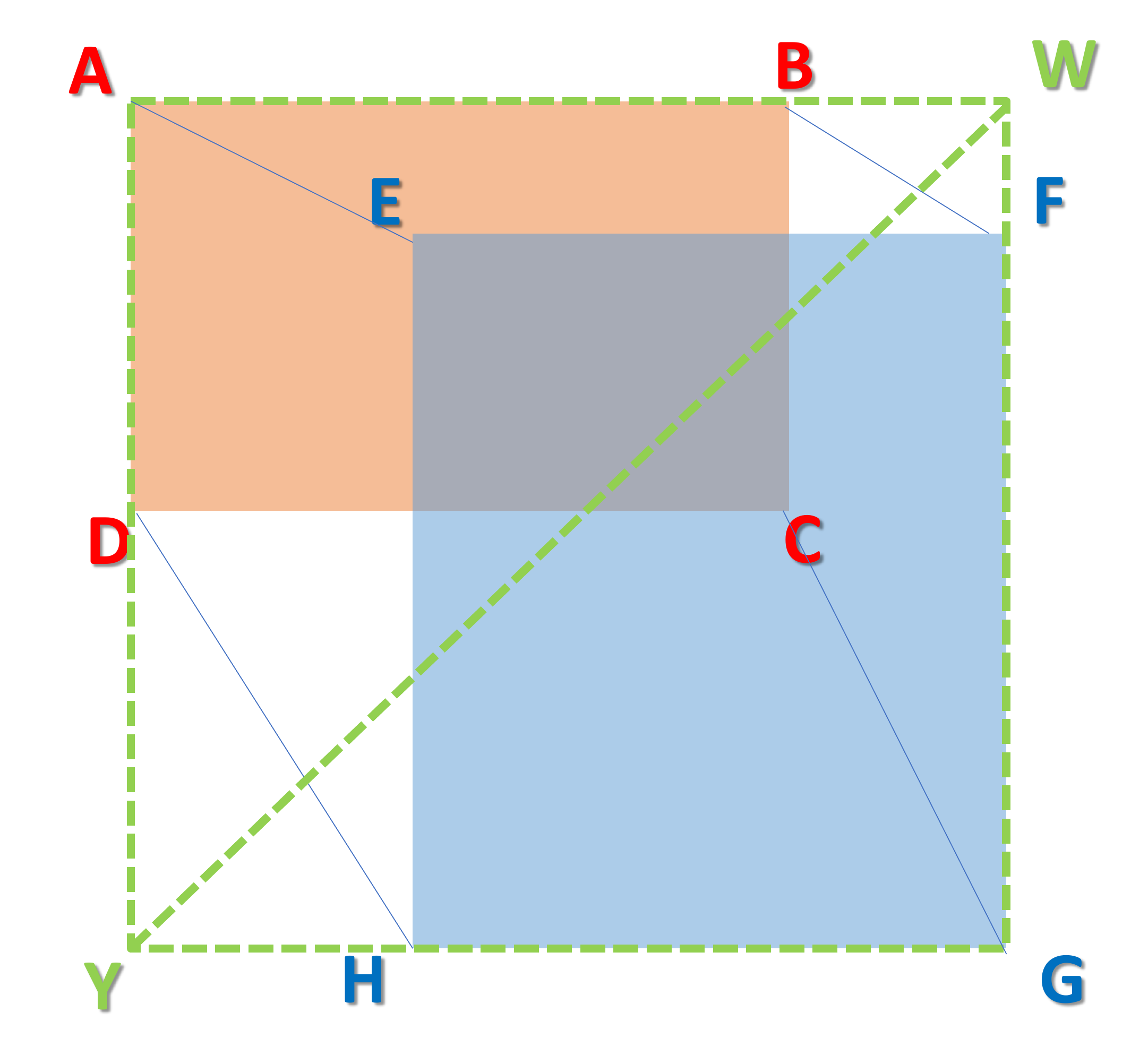}
\caption{Calculation of CDIoU. Minimum bounding rectangle (MBR) is the smallest rectangle that can contain RP and GT.}
\label{fig:label}
\end{figure}

\begin{equation}
C D I o U=I o U+\lambda(1-\operatorname{diou})
\end{equation}

Although the formula for CDIoU does not mention ``centroid distance" and ``aspect ratio",
the final calculation results reflect a measure of the degree of difference between
RP and GT. The higher the value of CDIoU,
the lower the degree of difference; the higher the value of CDIoU, the higher the similarity.

\begin{equation}
   \mathcal{L}_{C D I o U}=\mathcal{L}_{I o U_{s}}+\operatorname{diou}
\end{equation}

In order to cooperate with the calculation of CDIoU, this paper also proposes the CDIoU loss function.
By observing this formula, we can intuitively feel that after backpropagation,
the deep learning model tends to pull the four vertices of the region proposal toward the four vertices
of the ground truth until they overlap. For detailed information, see \textbf{Algorithm 1} and Figure 6.

CDIoU and CDIoU loss as a new metric have the following properties:
(1) \begin{math}0 \leq \text { diou }<1 \end{math}, \begin{math}\mathcal{L}_{I o U_{s}}\end{math} is the lower limit of \begin{math}\mathcal{L}_{C D I o U} \end{math}.
(2) \begin{math} \text { diou } \end{math} broadens the range of \begin{math}\mathcal{L}_{C D I o U} \end{math}. If \begin{math} \mathcal{L}_{I o U_{s}} = \mathcal{L}_{I o U} \end{math}, then \begin{math}0 \leq  \mathcal{L}_{C D I o U}<2 \end{math}, if \begin{math} \mathcal{L}_{I o U_{s}} = \mathcal{L}_{GI o U} \end{math}, then \begin{math}-1 \leq  \mathcal{L}_{C D I o U}<2 \end{math}.

\begin{figure}[h]
\centering
\includegraphics[scale=0.18]{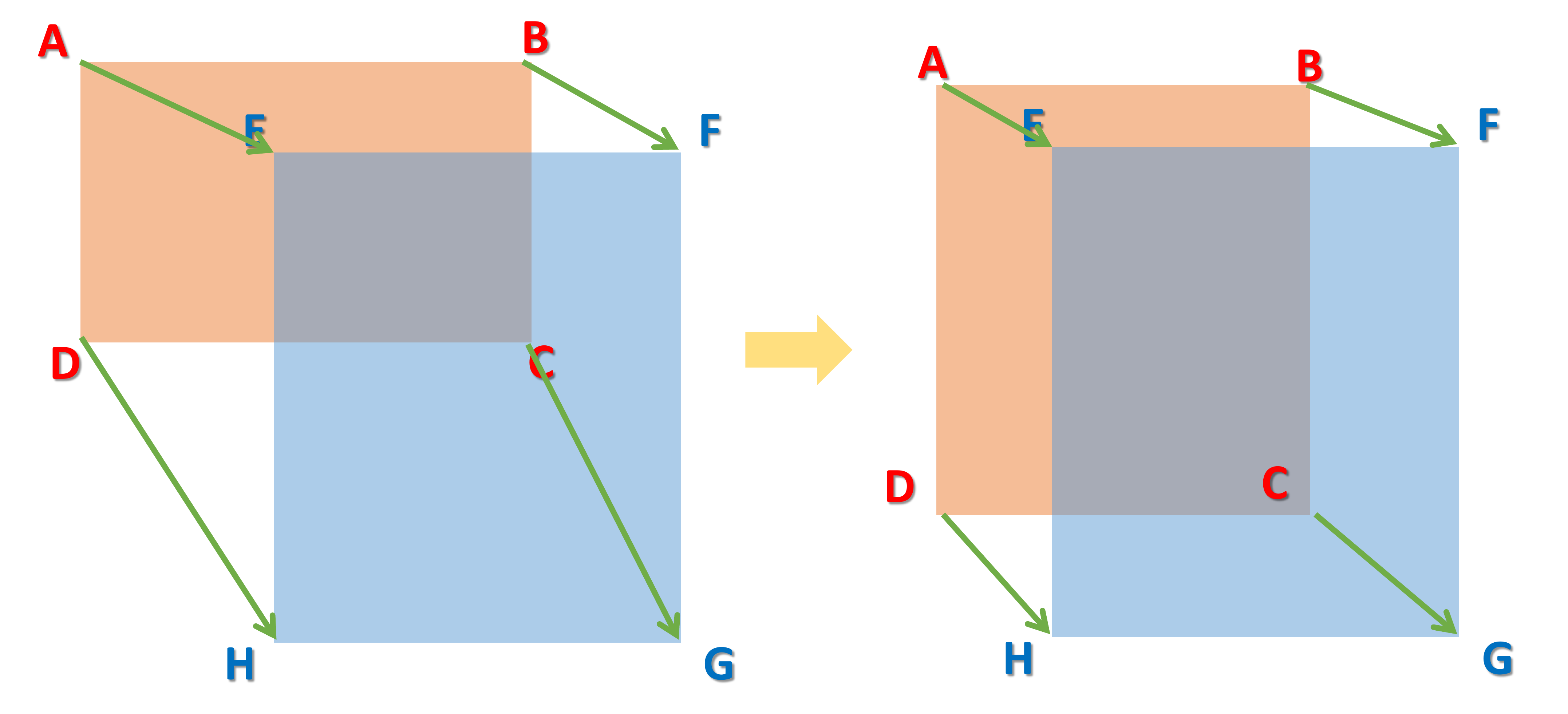}
\caption{Region proposal turning and variations by CDIoU loss}
\label{fig:label}
\end{figure}

\begin{algorithm}[htb]
  \caption{ CDIoU and CDIoU loss function }
  \label{alg:Framwork}
  \begin{algorithmic}[1]
    \Require
      RP for region proposal;
      GT for ground truth;
    \Ensure
      CDIoU and CDIoU loss;
    \State For RP and GT, find MBR;
    \label{code:fram:extract}
    \State compute $I o U=\frac{|RP \cap GT|}{|RP \cup GT|}$ ,
    $\operatorname{diou} =\frac{\|R P-G T\|_{2}}{4\mathrm{MBR}^{\prime} s \text { diagonal }}$;
    \label{code:fram:trainbase}
    \State compute $C D I o U=I o U+\lambda(1-\operatorname{diou})$;
    \label{code:fram:add}
    \State compute $\mathcal{L}_{C D I o U}=\mathcal{L}_{I o U_{s}}+\operatorname{diou}$, $\mathcal{L}_{I o U_{s}}$ could be $\mathcal{L}_{IoU}=-\ln(IoU)$, $\mathcal{L}_{IoU}=1-IoU$, $\mathcal{L}_{IoU}=1-IoU$ or $\mathcal{L}_{DIoU}$, $\mathcal{L}_{CIoU}$;
    \label{code:fram:classify}
  \end{algorithmic}
\end{algorithm}

\section{Experiments}

To ensure the rigidity and richness of the experiments,
we did a lot of training and testing on representative models,
such as Faster R-CNN, Cascade R-CNN\cite{cai2018cascade}, YOLOs and ATSS\cite{zhang2020bridging}.
Also, we try not to use
tricks and allow individual models to compare differences purely due to changes in IoUs or loss functions.

\subsection{Working environment and preparation}

The following experiments were conducted on MS coco 2017 dataset using two GeForce RTX 2080 Ti GPUs or two
Tesla V100 PCIe 32GB GPUs. All models under pytorch or tenserflow framework are standard models without using any tricks.

\subsection{CDIoU and CDIoU loss for object detection}

In order to verify the effectiveness of CDIoU and CDIoU loss in object detection,
experiments are designed and applied to numerous models in this paper.
These models encompass existing classical models and emerging models,
reflecting certain robustness and wide adaptability. We conduct several experiments to
study the robustness of the hyperparameter $\lambda$ in Table 1. Overall,
the only hyperparameter $\lambda=0.001$ is quite robust and the proposed
CDIoU can be nearly regarded as hyperparameter-free\footnote {Code is available in https://github.com/Alan-D-Chen/CDIoU-CDIoUloss}.

\begin{table}[]
\centering
\begin{tabular}{|c|c|c|c|c|c|}
\hline
$\lambda$                        & 1.0 & 0.1 & 0.01 & 0.001 & 0.0001 \\ \hline\hline
Faster R-CNN                   &35.6 &36.5 &37.0  &38.5   &  38.0  \\
ATSS\_R\_50\_FPN            &38.1 &39.0 &39.3  &39.5   &  38.9  \\ \hline
\end{tabular}
\caption{Analysis(\%) of different values of hyperparameter $\lambda$ on the \textbf{MS COCO} \texttt{val} set.}
\end{table}

% \usepackage{multirow}

\begin{table*}[h]
   \centering
   \begin{threeparttable}
   \begin{tabular}{|l|c|c|c|c|c|c|c|c|}
   \hline
   \multicolumn{1}{|c|}{\multirow{2}{*}{Models}}                            & \multicolumn{4}{c|}{test-dev(\%)} & \multicolumn{4}{c|}{val(\%)}                                                                                 \\ \cline{2-9}
   \multicolumn{1}{|c|}{}                                                               & AP     & APs    & APm    & APl    & AP    & APs   & APm   & APl                                                                              \\ \hline \hline
   Faster R-CNN\cite{2015Fast}                                                    & 36.0   & -      & -      & -      & 36.0  & -     & -     & -                                                                                                    \\
   YOLOv4\cite{bochkovskiy2020yolov4}                                      & 41.2   & 20.4   & 44.4   & 56.0   & -     & -     & -     & -                                                                                             \\
   RetinaNet-R101 (1024)\cite{2017Focal}                                    & 40.8   & 24.1   & 44.2   & 51.2   & -     & -     & -     & -                                                                                              \\
   ResNet-50 + NAS-FPN (1280@384)\cite{2019NAS}                  & 45.4   & -      & -      & -      & -     & -     & -     & -                                                                                              \\
   Detectron2 Mask R-CNN R101-FPN\cite{wu2019detectron2}    & -      & -      & -      & -      & 44.3  & -     & -     & -                                                                                              \\
   Cascade R-CNN\cite{cai2018cascade}                                       & 42.8   & 23.7   & 45.5   & 55.2   & -     & -     & -     & -                                                                                              \\
   FCOS\cite{2020FCOS}                                                              & 43.2    & 26.5   & 46.2   & 53.3   & -     & -     & -     & -                                                                                              \\
   ATSS\_R\_50\_FPN\_1x\cite{zhang2020bridging}                      & 39.2   & -      & -      & -      & 39.2  & -     & -     & -                                                                                              \\
   ATSS\_dcnv2\_R\_50\_FPN\_1x\cite{zhang2020bridging}          & 43.0   & -      & -      & -      & 43.0  & -     & -     & -                                                                                              \\
   ATSS\_R\_101\_FPN\_2x\cite{zhang2020bridging}                    & 43.6   & -      & -      & -      & 43.5  & -     & -     & -                                                                                              \\
   ATSS\_dcnv2\_R\_101\_FPN\_2x\cite{zhang2020bridging}        & 46.3   & -      & -      & -      & 46.1  & -     & -     & -                                                                                              \\
   ATSS\_X\_101\_32x8d\_FPN\_2x\cite{zhang2020bridging}        & 45.1   & -      & -      & -      & 44.8  & -     & -     & -                                                                                             \\
   ATSS\_dcnv2\_X\_101\_32x8d\_FPN\_2x\cite{zhang2020bridging}& 47.7   & -      & -      & -      & 47.7  & -     & -     & -                                                                                              \\
   ATSS\_dcnv2\_X\_101\_64x4d\_FPN\_2x\cite{zhang2020bridging}& 47.7   & -      & -      & -      & 47.7  & -     & -     & -                                                                                             \\ \hline\hline
   \multicolumn{9}{|c|}{Comparison test 1}                                                                                                                                                                                                                                     \\ \hline\hline
   ATSS\_R\_50\_FPN\_1x + IoU \& loss                                       & 38.6  &  20.7    &  37.4    &  45.7    & 38.5  & 21.0  & 33.7  & 50.9                                                                                        \\
   ATSS\_dcnv2\_R\_50\_FPN\_1x + IoU \& loss                           & 41.9  & 24.0    &  45.8     &  53.8    & 42.5  & 23.4  & 44.9  & 56.9                                                                                         \\
   ATSS\_dcnv2\_R\_101\_FPN\_2x + IoU \& loss                         & 45.8  &  25.9   &  48.6     &  56.9    & 45.7  & 28.5  & 49.7  & 60.0                                                                                         \\
   ATSS\_X\_101\_32x8d\_FPN\_2x + IoU \& loss                         & 44.5  &  26.8    &  47.2     &  53.2    & 44.7  & 27.7  & 46.1  & 55.9                                                                                         \\
   ATSS\_dcnv2\_X\_101\_32x8d\_FPN\_2x + IoU \& loss             & 46.8  &  27.9    &  49.7     &  58.7    & 47.0  & 29.7  & 50.1  & 60.2                                                                                         \\
   ATSS\_dcnv2\_X\_101\_32x8d\_FPN\_2x(MS) + IoU \& loss      & 49.6  &  31.2   &  50.5     &   60.3   & 49.9  & 32.7  & 50.9  &  61.4                                                                                         \\ \hline\hline
   \multicolumn{9}{|c|}{Comparison test 2}                                                                                                                                                                                                                                     \\ \hline\hline
   Faster R-CNN + CDIoU \& loss                                                   & \textbf{38.3}   & 17.3  & 38.0  & 54.4  & \textbf{38.5}  & 17.5  & 37.9  & 55.7                                                                                         \\
   YOLOv4 + CDIoU \& loss                                                            & \textbf{41.4}   & 20.4  & 46.1  & 55.8  & \textbf{41.9}  & 21.0  & 46.7  & 57.1                                                                                         \\
   RetinaNet-R101 (1024) + CDIoU \& loss                                      & \textbf{41.2}   & 22.5  & 43.1  & 50.9  & \textbf{41.5}  & 24.5  & 45.3  & 52.7                                                                                         \\
   ResNet-50 + NAS-FPN (1280@384) + CDIoU \& loss                   & \textbf{45.8}   & 22.1  & 48.0  & 65.1  & \textbf{46.0}  & 22.1  & 48.9  & 67.4                                                                                         \\
   Detectron2 Mask R-CNN R101-FPN + CDIoU \& loss                  & \textbf{45.0}   & 21.3  & 46.0  & 64.9  & \textbf{45.2}  & 21.9  & 45.7  & 66.5                                                                                         \\
   Cascade R-CNN + CDIoU \& loss                                                & \textbf{43.0} & 25.0    & 45.7  & 66.1  & \textbf{43.3}  & 25.3  & 46.1  & 67.2                                                                                         \\
   ATSS\_R\_50\_FPN\_1x + CDIoU \& loss                                     & \textbf{39.4}&  22.5    &  42.2     &  49.8    & \textbf{39.5}  & 23.0  & 34.4  & 52.1                                                                                         \\
   ATSS\_dcnv2\_R\_50\_FPN\_1x + CDIoU \& loss                         & \textbf{43.1} & 24.4    &  46.0     &  55.8    & \textbf{43.1}  & 25.4  & 46.9  & 57.8                                                                                         \\
   ATSS\_dcnv2\_R\_101\_FPN\_2x + CDIoU \& loss                       & \textbf{46.4} &  27.8   &  49.7     &  58.6    & \textbf{46.3}  & 29.3  & 50.4  & 61.2                                                                                         \\
   ATSS\_X\_101\_32x8d\_FPN\_2x + CDIoU \& loss                       & \textbf{45.2} &  27.8    &  48.3     &  55.2    & \textbf{45.1}  & 28.7  & 48.5  & 57.9                                                                                         \\
   ATSS\_dcnv2\_X\_101\_32x8d\_FPN\_2x + CDIoU \& loss           & \textbf{47.9} &  29.6    &  50.8     &  60.5    & \textbf{48.1}  & 30.6  & 51.7  & 62.7                                                                                        \\
   ATSS\_dcnv2\_X\_101\_32x8d\_FPN\_2x(MS) + CDIoU \& loss    & \textbf{50.7} &  33.2   &  52.6     &   62.7   & \textbf{50.9}  & 35.8  & 53.4  &  65.3                                                                                         \\ \hline
   \end{tabular}
   \caption{Detection models results with and without IoUs and IoU loss function on the \textbf{MS COCO} \texttt{val} and \texttt{test-dev} set.}
    \begin{tablenotes}
        \footnotesize
        \item[1] \textbf{MS} means multi-scale testing.
        \item[2] \textbf{+ CDIoU \& loss} means that this model uses CDIoU and CDIoU loss as evaluation-feedback module.
        \item[3] \textbf{+ IoU \& loss} means that this model uses IoU and IoU loss as evaluation-feedback module.
        \item[4] The original ATSS models use IoU and GIoU loss as evaluation-feedback module, and the original Faster R-CNN,YOLOv4,RetinaNet-R101,ResNet-50 + NAS-FPN,Detectron2 Mask R-CNN,Cascade R-CNN models use IoU and IoU loss or \textit{L1-smooth} as evaluation-feedback module.
        \item[5] FCOS model uses the SSC method, so there is no comparison test.
        \item[6] Bold fonts indicate the best performance.
        \item[7] \textit{1x} and \textit{2x} mean the model is trained for 90K and 180K iterations, respectively.
        \item[8] All results are obtained with a single model and without any test time data augmentation such as \textit{multi-scale, flipping} and etc..
        \item[9] \textit{dcnv2} denotes deformable convolutional networks v2.
      \end{tablenotes}
    \end{threeparttable}
   \end{table*}

CDIoU and CDIoU loss are universally adaptable, exhibiting differential performance gains on different models.
As shown in Table 2, we can see that the more complex the backbone structure is,
the less the CDIoU and CDIoU loss improvement is, while for the basic model,
the CDIoU and CDIoU loss improvement is more obvious. By using CDIoU and CDIoU loss,
the models in Table 2 improved AP by an average of 0.8\%.

\subsection{Ablation studies}

Faster R-CNN is a classical model, while ATSS is a new model recently. In Table 3, the evaluation system of all models is IoU, the most basic original one.
But in the loss function, we selects different calculation functions as feedback mechanism. \textit{FPS} represents the number of images that can be processed per second. CDIoU loss does not increase the amount of computation and barely improves \textit{FPS}.

\renewcommand\arraystretch{1.1}
\begin{table}[h]
   \centering
   %\begin{spacing}{1.35}  %调整表格行距
   \begin{threeparttable}
   \begin{tabular}{|c|c|c|c|c|c|}
   %\label{table2:}
   \hline
   IoUs loss                    & Model                                                & AP                     & FPS                   \\ \hline\hline
   L1-smooth                 & Faster R-CNN\cite{2017Faster}            & 36.0                   & 7.5                     \\
   IoU loss                     & Faster R-CNN                                      & 36.8                   & 7.7                     \\
   GIoU loss                   & Faster R-CNN                                     & 36.9                   & 8.5                     \\
   DIoU loss                   & Faster R-CNN                                     & 38.0                   & 7.9                     \\
   CIoU loss                   & Faster R-CNN                                      & 38.2                   & 6.3                    \\
   CDIoU loss                  & Faster R-CNN                                     & 38.5                   & 7.7                    \\ \hline\hline
   L1-smooth                 & ATSS\_R\_50\_FPN\_1x\cite{zhang2020bridging} & 37.5    & 11.1                    \\
   IoU loss                     & ATSS\_R\_50\_FPN\_1x                         & 38.0                   & 10.8                    \\
   GIoU loss                   & ATSS\_R\_50\_FPN\_1x                        & 39.2                   & 11.0                   \\
   DIoU loss                   & ATSS\_R\_50\_FPN\_1x                        & 39.0                   & 11.3                   \\
   CIoU loss                   & ATSS\_R\_50\_FPN\_1x                         & 39.2                   & 8.8                    \\
   CDIoU loss                 & ATSS\_R\_50\_FPN\_1x                        & 39.4                   & 11.2                   \\ \hline
   \end{tabular}
   %\end{spacing}
   \caption{Comparison of effects and running results(\%) of various IoU losses on the \textbf{MS COCO} \texttt{val} set.}
       \begin{tablenotes}
        \footnotesize
        \item[1] The default evaluation module IoUs is IoU in this table.
      \end{tablenotes}
    \end{threeparttable}
   \end{table}

As shown in Table 3, we can accurately see that CDIoU loss function can significantly improve the AP results by 0.2$\sim$1.9 $\%$
compared to other loss functions, and this effect is more obvious in traditional and basic models.

\begin{figure}[h]
\centering
\includegraphics[scale=0.38]{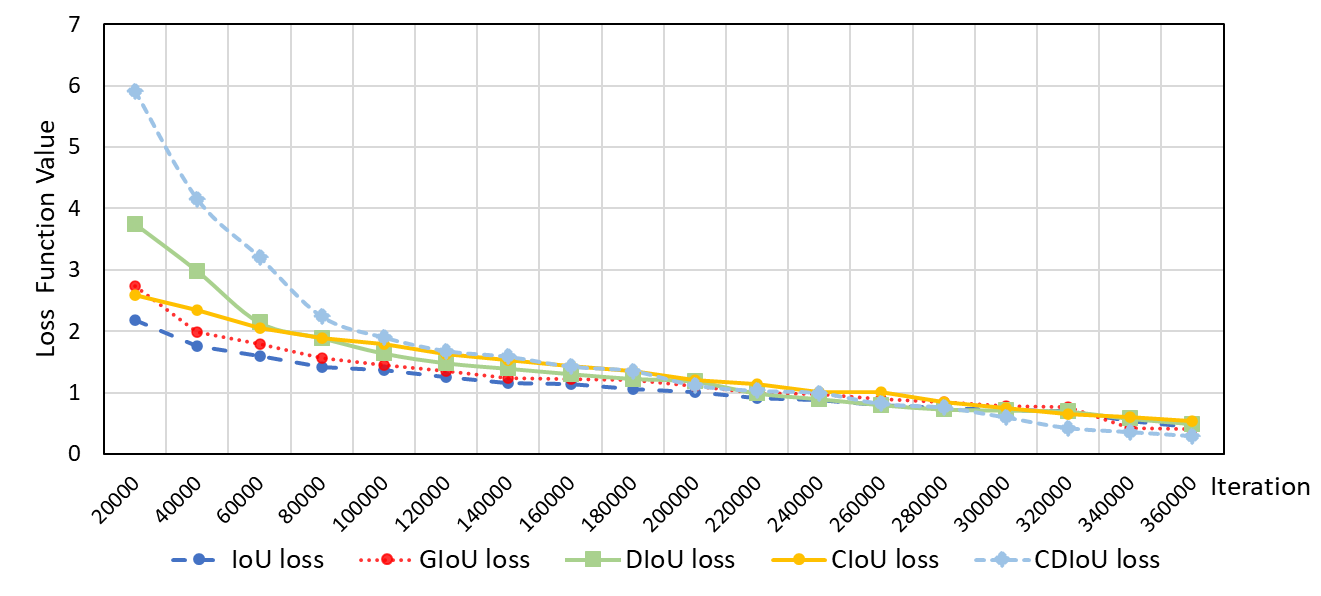}
\caption{Loss Function value of Faster R-CNN}
\label{fig:label}
\end{figure}

\begin{figure}[h]

\centering
\includegraphics[scale=0.36]{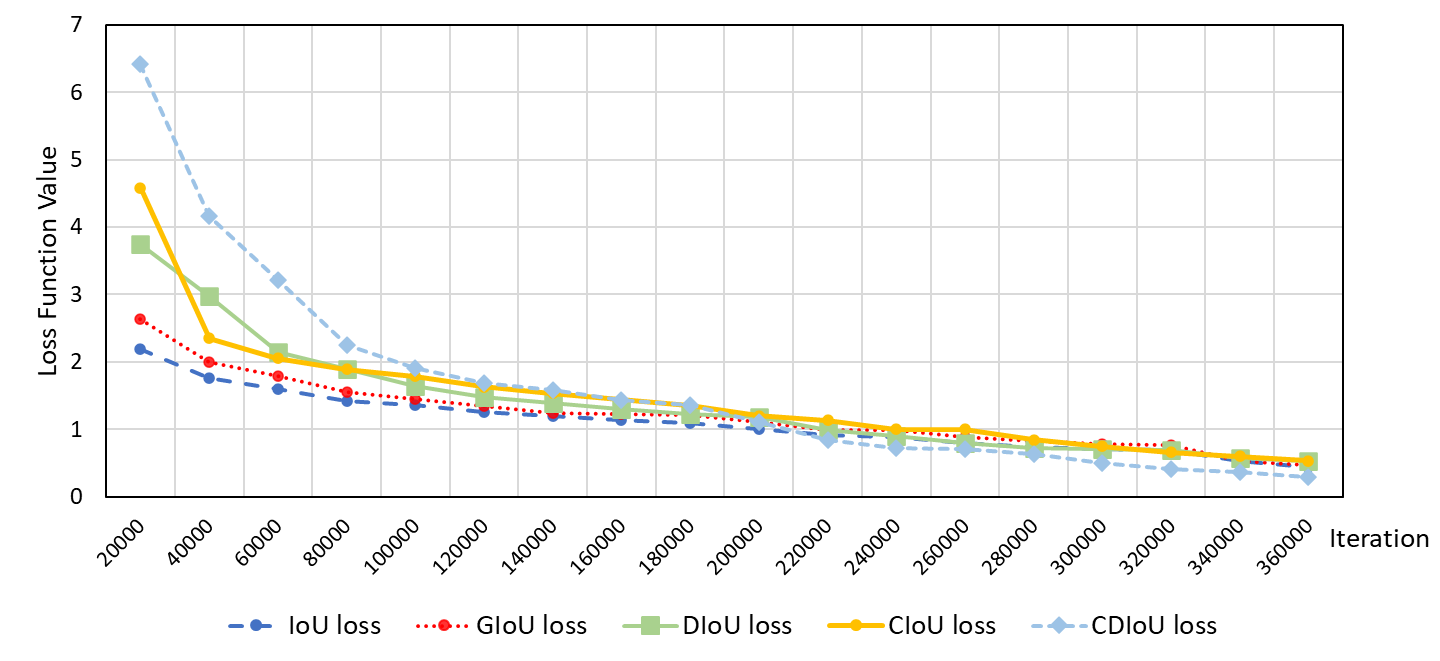}
\caption{Loss Function value of ATSS-R-50-FPN-1x}
\label{fig:label}
\end{figure}

\begin{figure}[h]

\centering
\includegraphics[scale=0.47]{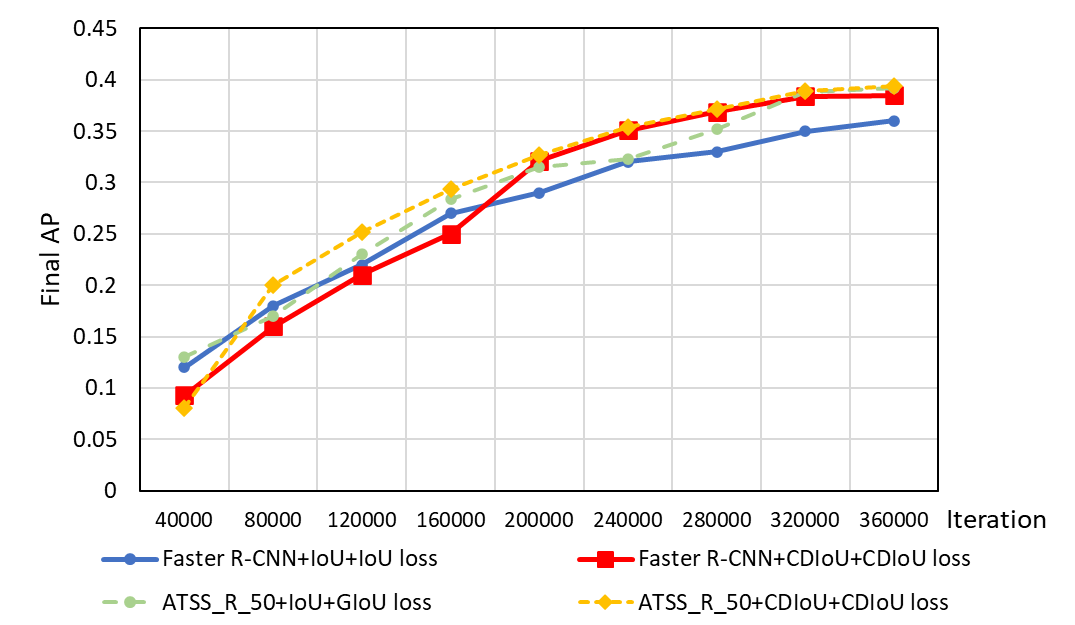}
\caption{The final AP of Faster R-CNN and ATSS-R-50-FPN-1x}
\label{fig:label}
\end{figure}

From Figure 7,8 , it is clear that the loss function values of both Faster R-CNN and
ATSS-R-50-FPN-1x models drop and reach convergence faster after using CDIoU and CDIoU loss function
compared to the other loss functions. The Figure 9 shows that the CDIoU and CDIoU loss functions
can help the models achieve higher AP values with fewer iterations. These results show that the CDIoU
and CDIoU loss functions have strong convergence and highlight their more accurate evaluation of region proposals.

In order to rigorously verify the effectiveness of CDIoU and CDIoU loss function
proposed in this paper, a large number of comparative experiments are designed to suggest supporting evidence.
We can learn from Table 4 that CDIoU loss can indeed achieve better results than IoU loss,
GIoU loss, DIoU loss, etc., while using the same IoU section.

\renewcommand\arraystretch{1.1}
\begin{table*}[h]
   \centering
   %\begin{spacing}{1.35}  %调整表格行距
   \begin{threeparttable}
   \begin{tabular}{|c|c|c|c|c|c|c|c|c|c|}
   \hline
   \multirow{2}{*}{Model}     & \multicolumn{2}{c|}{IoUs}             & \multicolumn{5}{c|}{IoU loss functions}                 & \multirow{2}{*}{AP}      \\ \cline{2-8}
                                          & GIoU      & CDIoU      & IoU loss & GIoU loss & DIoU loss & CIoU loss & CDIoU loss &                                   \\ \hline \hline
   Faster R-CNN                               & \checkmark&            &\checkmark&           &           &           &            &   36.9                                 \\
   Faster R-CNN                               & \checkmark&            &          &\checkmark &           &           &            &   37.3                                 \\
   Faster R-CNN                               & \checkmark&            &          &           & \checkmark&           &            &   38.0                                 \\
   Faster R-CNN                               & \checkmark&            &          &           &           &\checkmark &            &   38.2                                 \\
   Faster R-CNN                               & \checkmark&            &          &           &           &           & \checkmark &   38.3                                 \\ \hline
   Faster R-CNN                               &           &\checkmark  &\checkmark&           &           &           &            &   37.1                                 \\
   Faster R-CNN                               &           &\checkmark  &          &\checkmark &           &           &            &   37.3                                 \\
   Faster R-CNN                               &           &\checkmark  &          &           &\checkmark &           &            &   38.1                                 \\
   Faster R-CNN                               &           &\checkmark  &          &           &           & \checkmark&            &   38.3                                \\
   Faster R-CNN                               &           &\checkmark  &          &           &           &           &\checkmark  &   \textbf{38.5}                  \\ \hline\hline
   ATSS\_R\_50\_FPN\_1x                       &\checkmark &            &\checkmark&           &           &           &            &   38.0                           \\
   ATSS\_R\_50\_FPN\_1x                       &\checkmark &            &          &\checkmark &           &           &            &   39.3                           \\
   ATSS\_R\_50\_FPN\_1x                       &\checkmark &            &          &           & \checkmark&           &            &   39.0                           \\
   ATSS\_R\_50\_FPN\_1x                       &\checkmark &            &          &           &           &\checkmark &            &   39.2                           \\
   ATSS\_R\_50\_FPN\_1x                       &\checkmark &            &          &           &           &           & \checkmark &   39.4                           \\ \hline
   ATSS\_R\_50\_FPN\_1x                       &           &\checkmark  &\checkmark&           &           &           &            &   38.1                           \\
   ATSS\_R\_50\_FPN\_1x                       &           &\checkmark  &          &\checkmark &           &           &            &   39.3                           \\
   ATSS\_R\_50\_FPN\_1x                       &           &\checkmark  &          &           &\checkmark &           &            &   39.2                           \\
   ATSS\_R\_50\_FPN\_1x                       &           &\checkmark  &          &           &           &\checkmark &            &   39.2                           \\
   ATSS\_R\_50\_FPN\_1x                       &           &\checkmark  &          &           &           &           &\checkmark  &  \textbf{39.5}                   \\ \hline
   \end{tabular}
   \caption{Comparison of effects and running results(\%) of various IoUs and IoU loss functions on the \textbf{MS COCO} \texttt{val} set.}
       \begin{tablenotes}
        \footnotesize
        \item[1] For more information about results of evaluation module IoU, please refer to table 3.
      \end{tablenotes}
    \end{threeparttable}
   \end{table*}

By comparing the above experiments, we can observe that using different IoU modules with different IoU loss
functions yields different and thought-provoking results. Combining Table 3,4 , we can analyze that under the same IoU loss function condition,
using CDIoU module alone can improve the final result by 0.3 $\sim$ 1.8$\%$; under the same IoU module,
using CDIoU loss function alone can improve the final result by 0.2 $\sim$ 1.7$\%$.

At the same time, if IoU module and IoU loss function could be unified in the computational form (eg.GIoU + GIoU loss function and CDIoU + CDIoU loss function),
the final result would seem to be better than the sum of the results of using the two optimization schemes independently.

In order to verify the independence and validity of CDIoU and CDIoU loss function, we designs the following comparison test in Table 5,
using original IoU + IoU loss(or DIoU loss) function and CDIoU + CDIoU loss function in training and validation stages of the program respectively,
and finally comparing their AP results. We use CDIoU and CDIoU loss in the training and validation stages respectively,
which can improve the final AP. If we used CDIoU and CDIoU loss in both training and validation stages, we could get more significant improvement results.

\subsection{Analysis of experiments}

The improvement effect of CDIoU + CDIoU loss tends to decrease as the model is updated. First, as the backbone of the model deepens,
the model itself enhances the strength of feature extraction. Second,
the continuous improvement of FPN modules also optimizes the function of traditional evaluation systems.
The above two points offset the advantages of CDIoU and CDIoU loss compared with the traditional evaluation-feedback modules.

ATSS bridges the gap between anchor-based and anchor-free detection via adaptive training sample selection.
Comparison tests on ATSS exclude the essential interference between anchor-based and anchor-free detection.
In these tests, the interference of positive and negative sample generation is eliminated, which give tests based on ATSS more representativeness.

% \usepackage{multirow}
\renewcommand\arraystretch{1.1}
\begin{table*}[h]
\centering
%\begin{spacing}{1.35}  %调整表格行距
\begin{threeparttable}
\begin{tabular}{|c|c|c|c|c|c|c|c|}
\hline
\multirow{2}{*}{Model}       & \multicolumn{3}{c|}{Training}                                       & \multicolumn{2}{c|}{validation}                & \multirow{2}{*}{AP}       \\ \cline{2-6}
                                         & IoU+IoU loss & IoU+GIoU loss & CDIoU+CDIoU loss & Originals & CDIoU+CDIoU loss       &                                    \\ \hline \hline
Faster R-CNN                    & \checkmark   &               &                                         &\checkmark &                        &         36.8                             \\
Faster R-CNN                    & \checkmark   &               &                                         &           &  \checkmark            &         37.2                             \\ \hline
Faster R-CNN                    &                      &               &  \checkmark                     &\checkmark &                        &         37.7                             \\
Faster R-CNN                    &                      &               &  \checkmark                     &           &  \checkmark            &\textbf{38.5}                         \\ \hline\hline
ATSS\_R\_50\_FPN\_1x       &                      &  \checkmark   &                                & \checkmark&                        &         39.2                              \\
ATSS\_R\_50\_FPN\_1x       &                      &  \checkmark   &                                 &           &  \checkmark            &         39.3                             \\ \hline
ATSS\_R\_50\_FPN\_1x       &                      &               &    \checkmark                  & \checkmark&                        &         39.3                              \\
ATSS\_R\_50\_FPN\_1x       &                      &               &    \checkmark                   &           &  \checkmark            &\textbf{39.5}                         \\ \hline
\end{tabular}
\caption{Comparison of effects and running results(\%) in training and validation on the \textbf{MS COCO} \texttt{train} and \texttt{val} set.}
\begin{tablenotes}
        \footnotesize
        \item[1] \textit{Originals} means that original Faster R-CNN uses IoU+IoU loss as evaluation-feedback module and original ATSS\_R\_50\_FPN\_1x  uses IoU+GIoU loss as evaluation-feedback module.
      \end{tablenotes}
    \end{threeparttable}
\end{table*}

From Table 5 , we can clearly observe that replacing IoU module and loss function separately can improve the results of the original model,
and replacing IoU module and loss function at the same time also has a certain additive effect, achieving the synergy of ``one plus one is greater than two".
It lies on the calculation form consistency between evaluation system and feedback mechanism. The numerical fluctuations of the feedback mechanism reflect the differences of the evaluation system, which makes the evaluation-feedback module more targeted.

\subsection{Tips to improve performances}

In this experiment, we also found some tips to improve AP. These tips are particularly useful for some basic models in this paper.

\textbf{Floating learning rate}

It is a consensus that the learning rate decreases as the iterative process in the experiment. Further,
this paper proposes to check the loss every $\mathcal{K}$ iterations and increase the learning rate slightly, if the loss function does not decrease continuously.
In this way, the learning rate will decrease and float appropriately at regular intervals to promote the decrease of the loss function.

\begin{equation}
\begin{array}{ll}
l r=\left\{\begin{aligned}
1.05lr \textit{, Ls}<0 \\
lr       \textit{, Ls}>0
\end{aligned}\right. \\
\textit{Ls}=\operatorname{loss}_{i}-\operatorname{loss}_{i-k}
\end{array}
\end{equation}

\textbf{Automatic GT clustering analysis}

It is well known that AP can be effectively improved by performing cluster analysis on GT in the original dataset.
We adjust anchor sizes and aspect ratios parameters based on the results of this cluster analysis.
However, we do not know the number of clusters through the current approach.
The main solution is to keep trying the number of clusters $\mathcal{N}$,
and then judge by the final result AP. Obviously, this exhaustive method takes a lot of time.

In this paper, automatic GT clustering analysis is proposed, using \textit{K-means/PAM},
\textit{Hierarchical Clustering}, \textit{Spectral Clustering}, \textit{DBSCAN} and \textit{Mean-shift} methods respectively,
where \textit{DBSCAN} and \textit{Mean-shift} methods are able to obtain the number of clusters autonomously.
The above methods were evaluated using \textit{SSE} (sum of the squared errors), \textit{Silhouette Coefficient} and \textit{Calinski-Harabaz},
and then two recommended schemes are obtained. These recommended schemes include the number of clusters
and the central GT of each cluster. We obtained anchor information from the central GT
before executing the complex deep learning network, so that the experiments in this paper
are much more efficient.

\section{Conclusion}

In this paper, we propose that a good evaluation-feedback module should focus on
both evaluation system and feedback mechanism. An evaluation-feedback module
should meet 3 conditions (overlapping area,
the degree of difference and correlated with the loss function)
and 3 main roles (evaluation,ranking and feeding the gap).
Finally this paper proposes CDIoU and CDIoU loss,
a unified interrelated evaluation-feedback module without increasing the model running time and system memory.
Through a large number of experiments,
we can demonstrate that CDIoU and CDIoU loss can effectively improve
the performance of multiple object detection models without  increasing parameters or FLOPs in models.
By using CDIoU and CDIoU loss, the models improved their AP by an average of 0.8\%.
And there are a maximum AP performance improvement of 1.9\% and an average AP performance of 0.8\% improvement on MS COCO
dataset, comparing to traditional evaluation-feedback modules.
It lies on the calculation form consistency between evaluation system and feedback mechanism,
which makes the evaluation-feedback module more targeted.

\section*{Acknowledgements}
This research was supported in part by the National Natural Science
Foundation of China 61976158 and Grant Nos.~62006172.

{\small
\bibliographystyle{ieee_fullname}
\bibliography{egbib}
}

%\bibliography{egbib}

\end{document}